\documentclass{article}
\usepackage{spconf}
\usepackage{hyperref}
\usepackage{amsmath,array,graphicx}
\usepackage{url}
\usepackage{multirow}
\usepackage{hyperref}
\usepackage[table]{xcolor}
\usepackage{tabu}
\usepackage{color}

\newcolumntype{K}[1]{>{\centering\arraybackslash}p{#1}}

\title{Breast density classification with deep convolutional neural networks}

\name{
\begin{tabular}{c}Nan Wu${^1}$, Krzysztof J. Geras${^1}$, Yiqiu Shen${^1}$, Jingyi Su${^1}$, S. Gene Kim${^{3,4}}$, Eric Kim${^3}$, Stacey Wolfson${^3}$,\\
Linda Moy${^{3,4}}$ \& Kyunghyun Cho${^{1,2}}$\end{tabular}
}

\address{
${^1}$Center for Data Science, New York University\\
${^2}$Courant Institute of Mathematical Sciences, New York University\\
${^3}$Center for Biomedical Imaging, Radiology, NYU School of Medicine\\
${^4}$Perlmutter Cancer Center, NYU Langone Medical Center
}

\begin{document}

\maketitle

\begin{abstract}
Breast density classification is an essential part of breast cancer screening. Although a lot of prior work considered this problem as a task for learning algorithms, to our knowledge, all of them used small and not clinically realistic data both for training and evaluation of their models. In this work, we explore the limits of this task with a data set coming from over 200,000 breast cancer screening exams. We use this data to train and evaluate a strong convolutional neural network classifier. In a reader study, we find that our model can perform this task comparably to a human expert.
\end{abstract}


\section{Introduction}

Although convolutional neural networks (CNNs) are highly successful in a variety of applications \cite{lecun2015deep}, they have received relatively little attention in medical image analysis until recently. This has been primarily due to the lack of availability of large public data sets. One of the significant areas for development in medical image analysis is breast cancer screening. Even though performing full diagnosis by the means of a neural network remains a challenge \cite{high_resolution}, elements of achieving this wider goal are feasible with current state of the art methods. In this paper, we explore one of them, namely breast density classification.

Mammographic density reflects the composition of fibroglandular and fat tissue of a breast as seen on a mammogram. In clinical practice in the United States, breast density is qualitatively categorized into four types: a) almost entirely fatty, b) scattered areas of fibroglandular density, c) heterogeneously dense and d) extremely dense. The last two categories are considered ``dense'' \cite{d2013acr}. Dense breast tissue is common and is the typical fibroglandular breast tissue pattern in young women. About 50\% of women over the age of 40 years have dense breasts. After menopause the breasts tend to contain more fat as the glands involute. Dense breast tissue reduces the effectiveness of mammography because it has a ``masking effect'' and will hide an underlying tumor. Studies also consistently show an increased risk of developing breast cancer in women with high mammographic density compared with women with low mammographic density.

Masking of cancer by dense tissue has become a political issue with women requesting supplemental tests if they have dense breasts. There are currently states in the US that mandate women receive notification about breast density with their mammography results. Due to significant variability in the radiologist' assessment of breast density, computer methods have been developed to improve consistency. One of them, Cumulus, is a software program requiring manual input to outline and measure the area of breast tissue relative to overall breast area \cite{byng1998analysis, bergbreast}. Recently, automated programs have been developed that measure percent density as a function of area or volume. Several automated density programs have demonstrated high reproducibility \cite{alonzo2015reliability} and correlation with volumetric density as measured by MRI \cite{gubern2014volumetric}. However, these commercially available products do not involve any learning, therefore they lack the flexibility and robustness of learning models. On the other hand, all learning models in literature were trained and tested with small amounts of data. Hence there still is a need to have a precise automated assessment of breast density based on learning directly from data.

\section{Data}

We used a clinically realistic data set of over 200,000 screening mammography exams, each containing at least four images corresponding to the standard four views used in screening mammography \cite{high_resolution}. For the purposes of this research, we supplemented this data with labels corresponding to breast density, which we automatically extracted from the textual reports associated with the exams in our data set. A small number of exams in our original data did not have the information on breast density in the corresponding textual report. We excluded 519 such exams from the data set, which left us with the total of 201,179 exams, containing 19,939 class 0, 85,665 class 1, 83,852 class 2 and 11,723 class 3 exams. Interestingly, analysis of our data confirms that women who were assigned overall BI-RADS 0 (``incomplete'') label in their screening mammography tend to have denser breasts than the ones who were assigned BI-RADS 1 (``normal'') and BI-RADS 2 (``benign'') labels (cf. \autoref{tab:distribution}).

\begin{figure}[h!]
\begin{center}
\begin{tabular}{c c}
\includegraphics[width=0.2\textwidth, trim = 0mm 0mm 0mm 0mm, clip]{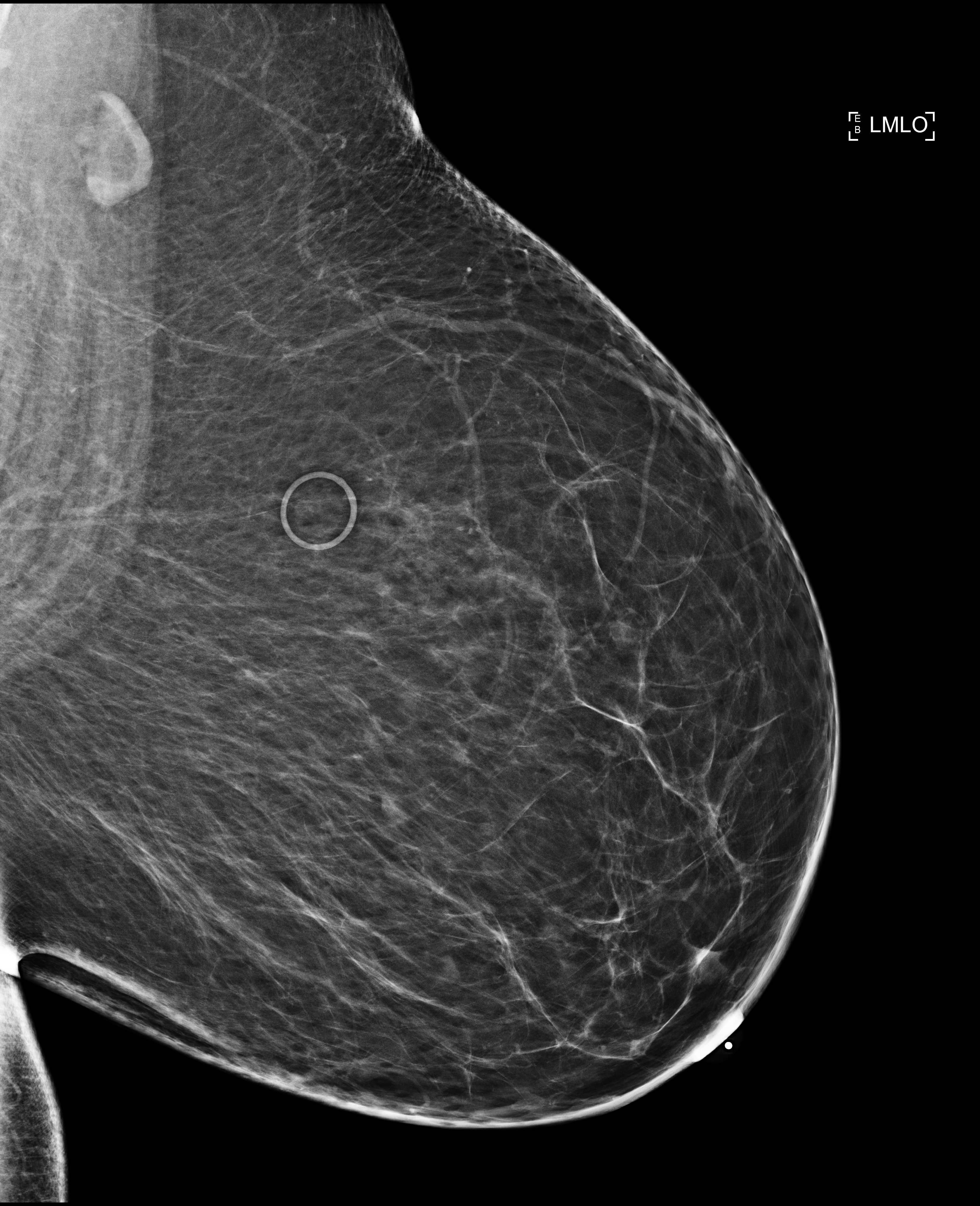} & \includegraphics[width=0.2\textwidth, trim = 0mm 0mm 0mm 0mm, clip]{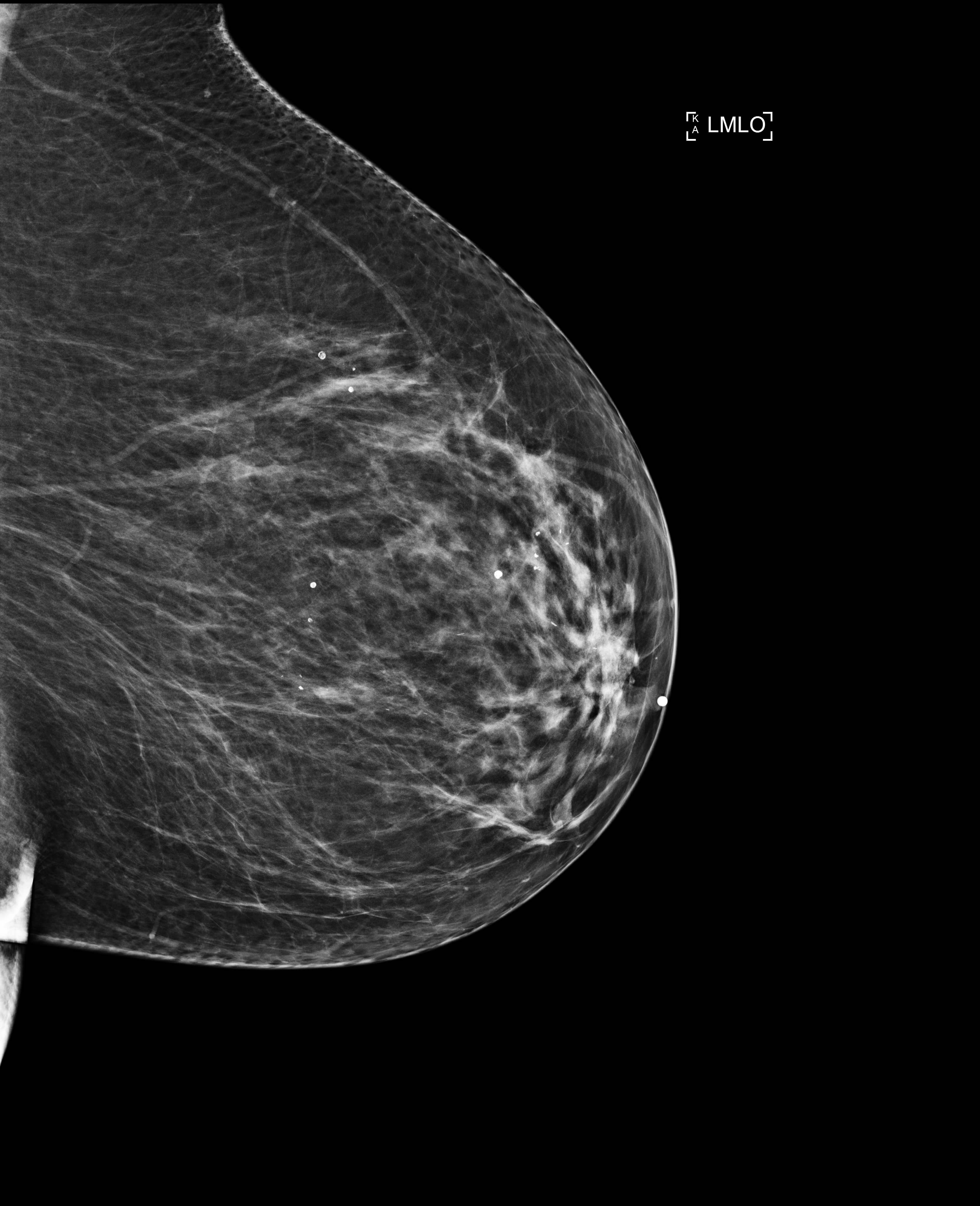}\\
almost entirely fatty (0) & \shortstack{scattered areas of\\fibroglandular density (1)}\\
\vspace{-0.3cm}\\
\includegraphics[width=0.2\textwidth, trim = 0mm 0mm 0mm 15mm, clip]{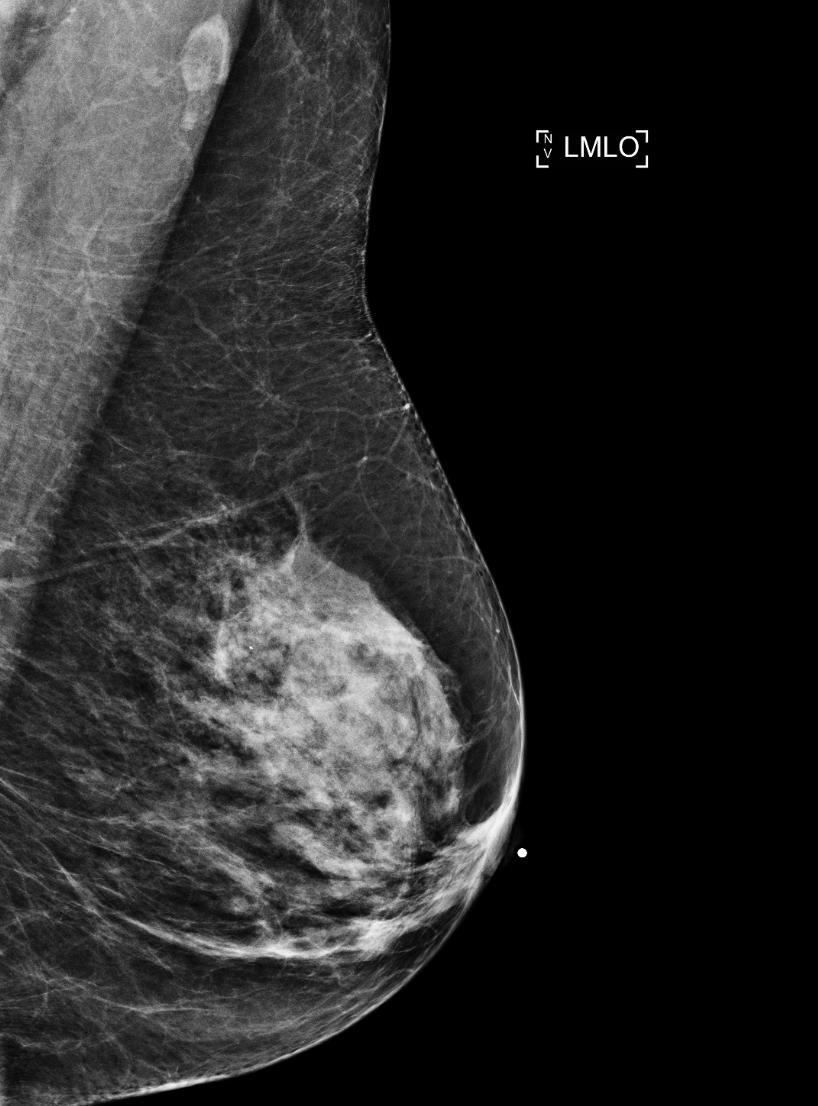} & \includegraphics[width=0.2\textwidth, trim = 0mm 0mm 0mm 0mm, clip]{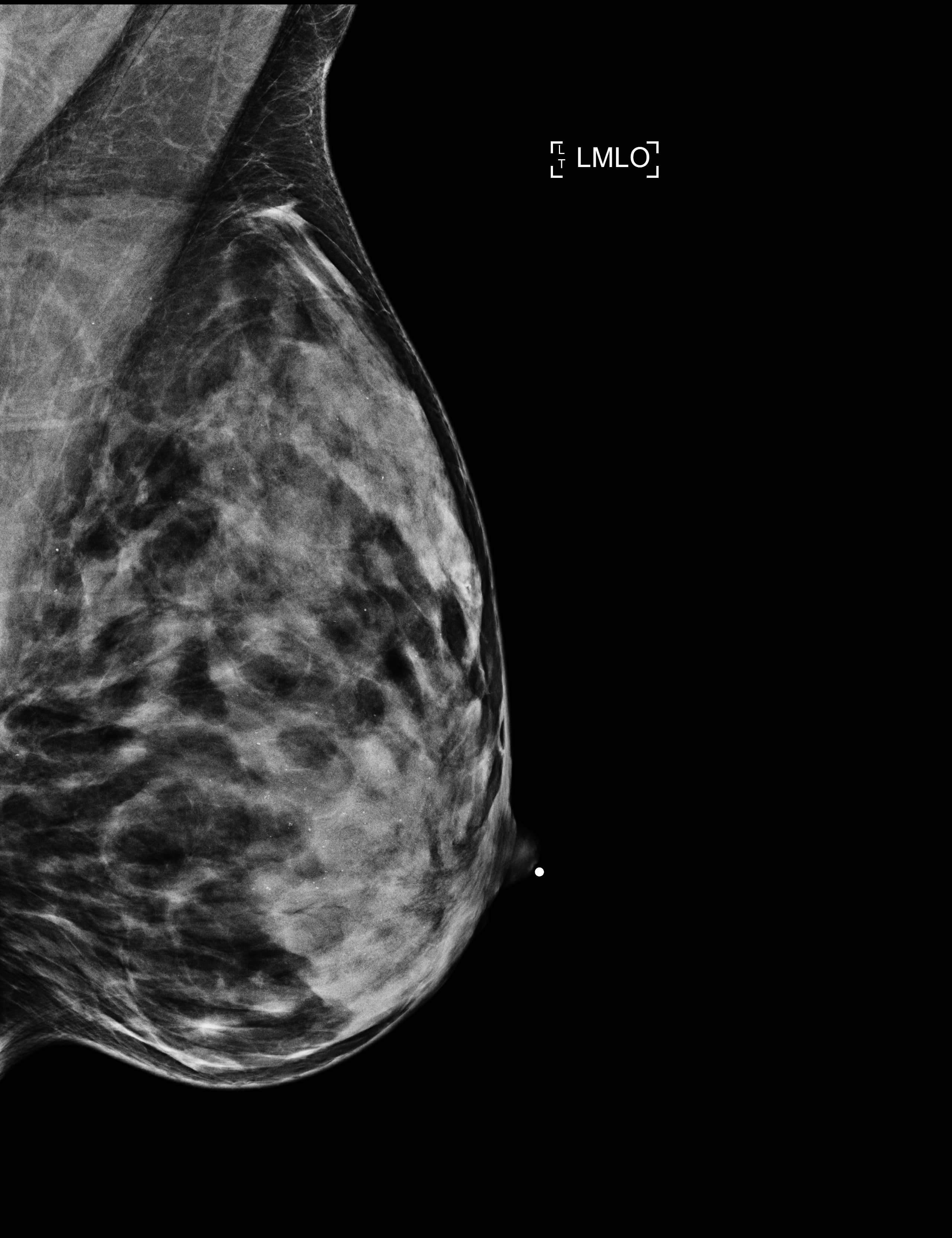} \\
heterogeneously dense (2) & extremely dense (3)
\end{tabular}
\end{center}
\vspace{-4mm}
\caption{Examples of the four breast density classes.}
\label{fig:examples}
\end{figure}

\begin{table}[h!]
\centering
\caption{Distribution of labels in our data set. The numbers in the bottom row are numbers of exams falling into different breast density categories. The numbers in the rightmost column are the numbers of exams falling into different overall BI-RADS classes.}
\label{tab:distribution}
\begin{tabular}{c c | c | c | c | c | c}
\cline{3-6}
 &  & \multicolumn{4}{c|}
 {breast density category} & \multicolumn{1}{!{\vrule width 1.7pt}c!{\vrule width 0 pt}}{ } \\ \cline{3-6}
 \multicolumn{1}{l}{} &  & 0 & 1 & 2& 3 & \multicolumn{1}{!{\vrule width 1.7pt}c!{\vrule width 0pt}}{ }\\ \hline
\multicolumn{1}{|c|}{\multirow{3}{*}{\rotatebox[origin=c]{90}{\small{BI-RADS}}}} & 0 & 1702 & 9607 & 12656 & 1839 & \multicolumn{1}{!{\vrule width 1.7pt}c!{\vrule width 0.1pt}} {25804} \\ \cline{2-7} 
\multicolumn{1}{|l|}{} & 1 & 9803 & 40060 & 37167 & 5157 & \multicolumn{1}{!{\vrule width 1.7pt}c!{\vrule width 0.1pt}} {92187}  \\ \cline{2-7} 
\multicolumn{1}{|l|}{} & 2 & 8434 & 35998 & 34029 & 4727 & \multicolumn{1}{!{\vrule width 1.7pt}c!{\vrule width 0.1pt}} {83188}  \\  \noalign{\hrule height 1.7pt}
 &  & 19939 & 85665 & 83852 & 11723 &\multicolumn{1}{!{\vrule width 1.7pt}c!{\vrule width 0pt}} {}  \\ \cline{3-6}
\end{tabular}
\end{table}

\section{Models}

\subsection{Baselines}
The most common method to perform the task of breast density prediction in literature is training a classifier with features based on histograms of pixel intensity in the image \cite{karssemeijer1998automated}. This simple method is surprisingly effective. The reason why it works can be easily understood. The difference between pixel intensity occurs because mammograms with a predominance of fat appear darker than the ones with a fibroglandular tissue. This is because this type of tissue absorbs much of the radiation whereas the adipose tissue allows the radiation to get through more easily.

In this work, we used such a model as a baseline. We took pixel intensity histograms as features and used softmax regression as a classifier.
For each of the four standard views (L-CC, R-CC, L-MLO and R-MLO) separately, we split intensity values into a set of predefined bins, we normalized them such that frequencies of different bins sum to one and then concatenate such feature vectors.
Additionally, to make this model more flexible, we also trained a version of it with an extra hidden layer with 100 hidden units between the input and the softmax regression layer. The hidden layer used rectifier linear function as an activation function.

\subsection{Deep convolutional neural network}

\begin{figure}[h!]
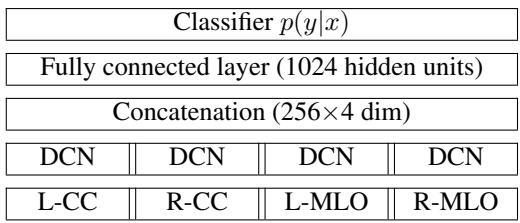

\begin{center}
\begin{tabular}{| K{1.2cm} || K{1.2cm} || K{1.2cm} || K{1.2cm} |}
\cline{1-4}
\multicolumn{4}{|c|}{Classifier $p(y|x)$} \\
\cline{1-4}
\multicolumn{4}{c}{} \\[-7pt]
\cline{1-4}
\multicolumn{4}{|c|}{Fully connected layer (1024 hidden units)} \\
\cline{1-4}
\multicolumn{4}{c}{} \\[-7pt]
\cline{1-4}
\multicolumn{4}{|c|}{Concatenation (256$\times$4 dim)} \\
\cline{1-4}
\multicolumn{4}{c}{} \\[-7pt]
\cline{1-1} \cline{2-2} \cline{3-3} \cline{4-4}
DCN & DCN & DCN & DCN \\
\cline{1-1} \cline{2-2} \cline{3-3} \cline{4-4}
\multicolumn{4}{c}{} \\[-7pt]
\cline{1-1} \cline{2-2} \cline{3-3} \cline{4-4}
L-CC & R-CC & L-MLO & R-MLO \\
\cline{1-1} \cline{2-2} \cline{3-3} \cline{4-4}
\end{tabular}
\end{center}
\vspace{-3mm}
\caption{An overview of the structure of the convolutional neural network used in our experiments. DCN in the figure above denotes a series of convolutional and pooling layers. L-CC, R-CC, L-MLO and R-MLO refer to inputs corresponding to the four standard views in screening mammography.}
\label{fig:architecture}
\end{figure}

We used a multi-column deep convolutional neural network of an architecture loosely inspired by the earlier work of Simonyan et al. \cite{simonyan2014very}. The input to the network is four $2600\times2000$ images corresponding to the standard views used in screening mammography. It is very similar to the architecture in \cite{high_resolution} with the exception of the number of the outputs in the softmax layer, since the breast density classification is a four-way classification problem (and not a three-way classification problem as the overall BI-RADS classification). We keep the same values of the hyperparameters too. An overview of our architecture is shown in \autoref{fig:architecture}.

\section{Experiments}

We sort the patients according to the date of their latest exam and divide them into training (first 80\%), validation (next 10\%) and test (last 10\%) sets. For the test phase, we only keep the most recent exam for each patient. This way of partitioning the data allows us to estimate performance of our classifiers on future data accurately. During experimentation, we follow the experimental protocol in \cite{high_resolution}, otherwise we state it explicitly when we deviate from it. In all experiments, we use data augmentation only during training. During validation and test phases all augmentations are off. For all models, we picked the best epoch according to the accuracy on the validation set.

In the baseline model based on histogram features, we use the Adam algorithm with the initial learning rate of $10^{-3}$. We tuned the number of bins of pixel intensity histogram, using the validation data to select between 10, 20, 50 and 100.

\subsection{Evaluation metrics}

Our primary metric in this work is the standard classification accuracy. As the levels of breast density correspond to relative increases in the amount of fibroglandular tissue, two consecutive labels can be confused even by an experienced radiologist. This is why we also considered \emph{top-k} accuracy. In this metric we consider a prediction is correct if the ground truth is among the $k$ most likely labels predicted. Top-k error is currently a popular performance measure on large scale image classification benchmarks such as ImageNet and Places \cite{Lapin2017AnalysisAO}. Additionally, we also considered accuracy only between the two superclasses: ``dense'' (classes 2 and 3) versus ``not dense'' (classes 0 and 1). 

Secondly, we evaluate our models with respect to the area under the ROC curve (AUC), which is widely used for measuring the predictive accuracy of binary classification models. This metric indicates the relation between the true positive rate and false positive rate when varying the classification threshold. As AUC can only be computed for binary classification, we compute AUCs for all four binary problems of distinguishing between one of the density categories and the rest of the density categories, and then take the macro average, abbreviated as macAUC.

\subsection{Impact of the size of the data set}
\begin{figure}[h!]
\begin{center}
\vspace{-2mm}
\includegraphics[width=0.75\linewidth, trim = 0mm 8mm 0mm 8mm, clip]{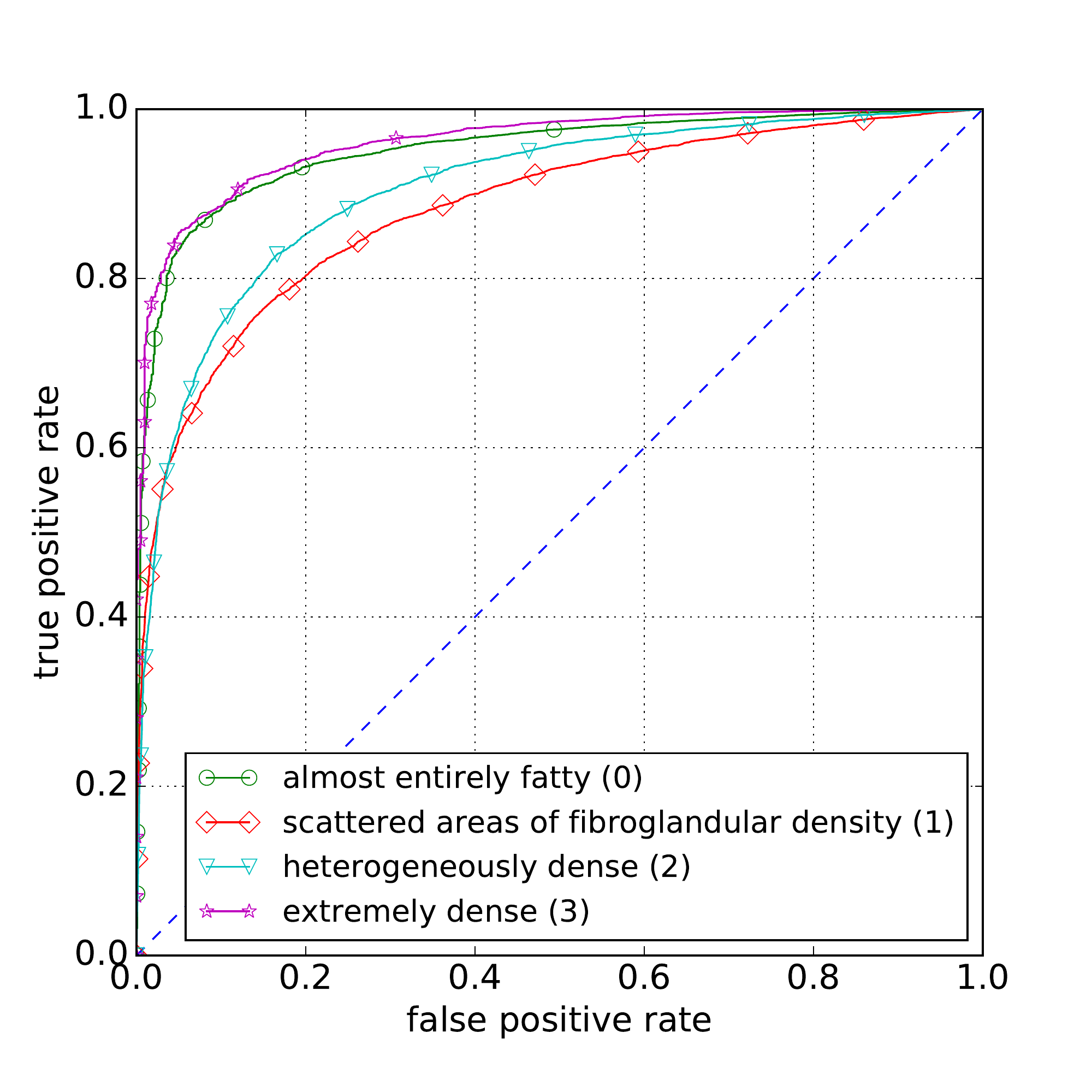}
\vspace{-2mm}
\caption{ROC curves for all four classes. The classes 1 and 2 are the hardest for a neural network to distinguish from the rest. The AUC values are 0.955, 0.888, 0.907, 0.960 for classes 0, 1, 2, 3 respectively.}
\label{fig:ROC}
\end{center}
\vspace{-4mm}
\end{figure}

To explore the effect of data set scale, we trained separate networks on training sets of different sizes; 100\%, 10\% and 1\% of the original training set. The results are shown in \autoref{tab:performance}. Interestingly, even though training with more data increases performance in all metrics, the difference is not large. In \autoref{fig:ROC} we show the ROC curves for the model trained with 100\% of the data. 

\begin{table}[h!]
\centering
\caption{Performance of our CNNs. The * symbol in the leftmost column indicates that a model was initialized using weights of a previously trained overall BI-RADS classifier (cf. \autoref{sec:transfer}).}
\label{tab:performance}
\resizebox{0.48\textwidth}{!}{
\begin{tabu}{| l | l | l | l | l | l |}
\hline
\multicolumn{1}{ | l |[2pt]}{data}  & \multicolumn{1}{ | l |[2pt]}{macAUC} & top-1 & top-2 & \multicolumn{1}{ | l |[2pt]}{top-3} & superclass \\ \tabucline[2pt]{1-6}
\multicolumn{1}{ | l |[2pt]}{1\% }  & \multicolumn{1}{ | l |[2pt]}{0.888}  & 0.729 & 0.967 & \multicolumn{1}{ | l |[2pt]}{0.998} & 0.849\\
\multicolumn{1}{ | l |[2pt]}{10\%}  & \multicolumn{1}{ | l |[2pt]}{0.907}  & 0.745 & 0.976 & \multicolumn{1}{ | l |[2pt]}{0.999} & 0.856\\
\multicolumn{1}{ | l |[2pt]}{100\%} & \multicolumn{1}{ | l |[2pt]}{0.916}  & 0.767 & 0.982 & \multicolumn{1}{ | l |[2pt]}{0.999} & 0.865\\ \tabucline[2pt]{1-6}
\multicolumn{1}{ | l |[2pt]}{*1\%}  & \multicolumn{1}{ | l |[2pt]}{0.892}  & 0.733 & 0.974 & \multicolumn{1}{ | l |[2pt]}{0.998} & 0.848\\
\multicolumn{1}{ | l |[2pt]}{*10\%} & \multicolumn{1}{ | l |[2pt]}{0.909}  & 0.753 & 0.980 & \multicolumn{1}{ | l |[2pt]}{0.998} & 0.856\\ \hline

\end{tabu}
}
\end{table}

\subsection{Transferring knowledge from BIRADS classifier}
\label{sec:transfer}

Transfer learning aims to transfer knowledge between related source and target domains \cite{pan2010survey}. In computer vision, examples of transfer learning include \cite{aytar2011tabula, tommasi2010safety, imagenet_transfer}. The main idea of this technique is to overcome the deficit of training samples by adapting strong classifiers trained for another, related but not identical, task. Considering the amount of parameters in the CNN and the correlation between breast density and overall BI-RADS (cf. \autoref{tab:distribution}), we applied the idea of transfer learning to accelerate learning of our breast density prediction network. To achieve that, we use the weights of our model previously trained for breast cancer screening \cite{high_resolution} to initialize the parameters of the network trained for breast density prediction. The two networks have an identical architecture, with the exception of the softmax layer. This layer of the network trained for breast density prediction is initialized randomly using the recipe from \cite{pmlr-v9-glorot10a}.

The models trained with such initialization perform better than their counterparts, trained from scratch in almost all metrics (cf. \autoref{tab:performance}), however, only by a small margin. Intriguingly, models initialized with parameters of a previously trained overall BI-RADS classifier achieve the best performance in much fewer numbers of training epochs: 20 instead of 50 when using 1\% of the original training data 15 instead of 25 when using 10\% of the original training data.

\subsection{Baseline results}
Finally, we trained the baselines based on histograms of pixel intensity.
Both versions of the baseline model are trained with 10\% of the original training set. The best baseline model without the hidden layer is the one with 20 bins. It achieved 0.832 of macAUC, 67.9\% of top-1 accuracy, 90.9\% of top-2 accuracy, 99.4\% of top-3 accuracy and 81.1\% in distinguishing between the ``dense'' and ``not dense'' superclasses. The model using 10 bins is the best one for the version with one hidden layer. It achieved 0.842 of macAUC, 69.4\% of top-1 accuracy, 90.8\% of top-2 accuracy, 99.2\% of top-3 accuracy and 82.5\% accuracy for superclass classification. 

\section{Comparison to human performance}

To understand what the limit of performance possible to achieve on this task is, we conducted a reader study with human experts with different levels of experience. The three participants in our reader study were: a medical student (S), a radiology resident (R) and an attending radiologist (A). The experts were all shown the same 100 exams randomly drawn from the test set, each with at least four images corresponding to the standard views used in screening mammography. For each exam, the experts were asked to rank the breast density classes from the most likely to the least likely according to their judgement.
The results of this experiment are shown in \autoref{tab:consistency}. Additionally, we computed analogous values with only two classes instead of the original four: dense breasts (original classes 2 and 3) and not dense breasts (original classes 0 and 1). The results of this experiment are shown in \autoref{tab:consistency_2}.

Both human experts and learning models achieve a fair agreement with the labels in the data. Note that the agreement between the predictions of our model and the labels in the data are of similar magnitude to the agreement between the humans themselves.  

We also compared our best CNN model to an average of the predictions of human experts. We did that by treating predictions of experts as one-hot vectors and averaging them. In this experiment the humans achieved macAUC of 0.892 (class 0: 0.960, class 1: 0.812, class 2: 0.807 and class 3: 0.990), while the CNN achieved macAUC of 0.934 (class 0: 0.971, class 1: 0.859, class 2: 0.905 and class 3: 1.000).

\begin{table}[h!]
\caption{Agreement (Cohen's kappa) in choosing the most likely class between different readers (S, R, A), our neural network (N), our baseline (H) and labels in the data set (L).}
\label{tab:consistency}
\vspace{-2mm}
\begin{center}
\footnotesize
\begin{tabu}{| c | [2pt]c | c | c | c | c | c |}
\cline{2-7}
\multicolumn{1}{c|}{} & L & N & H & S & R & A \\ \tabucline[2pt]{2-7} \hline
L & \cellcolor{gray!25} & 0.61 & 0.39 & 0.41 & 0.55 & 0.39\\ \hline
N & \cellcolor{gray!25} & \cellcolor{gray!25} & 0.58 & 0.53 & 0.60 & 0.48 \\ \hline
H & \cellcolor{gray!25} & \cellcolor{gray!25} & \cellcolor{gray!25} & 0.28 & 0.37 & 0.34 \\ \hline
S & \cellcolor{gray!25} & \cellcolor{gray!25} & \cellcolor{gray!25} & \cellcolor{gray!25} & 0.65 & 0.48\\ \hline
R & \cellcolor{gray!25} & \cellcolor{gray!25} & \cellcolor{gray!25} & \cellcolor{gray!25} & \cellcolor{gray!25} & 0.43\\ \hline
\end{tabu}
\end{center}
\vspace{-4mm}
\end{table}

\begin{table}[h!]
\caption{Agreement (Cohen's kappa) in distinguishing between dense breasts (classes 2 and 3) and not dense (classes 0 and 1) between different readers (S, R, A), our neural network (N), our baseline (H) and labels in the data set (L).}
\label{tab:consistency_2}
\vspace{-2mm}
\begin{center}
\footnotesize
\begin{tabu}{| c | [2pt]c | c | c | c | c | c |}
\cline{2-7}
\multicolumn{1}{c|}{} & L & N & H & S & R & A \\ \tabucline[2pt]{2-7} \hline
L & \cellcolor{gray!25} & 0.65 & 0.50 & 0.50 & 0.73 & 0.46\\ \hline
N & \cellcolor{gray!25} & \cellcolor{gray!25} & 0.72 & 0.62  & 0.83 & 0.57 \\ \hline
H & \cellcolor{gray!25} & \cellcolor{gray!25} & \cellcolor{gray!25} & 0.48 & 0.69 & 0.48\\ \hline
S & \cellcolor{gray!25} & \cellcolor{gray!25} & \cellcolor{gray!25} & \cellcolor{gray!25} & 0.69 & 0.64\\ \hline
R & \cellcolor{gray!25} & \cellcolor{gray!25} & \cellcolor{gray!25} & \cellcolor{gray!25} & \cellcolor{gray!25} & 0.60\\ \hline
\end{tabu}
\end{center}
\vspace{-3mm}
\end{table}

\section{Related work}

Many previous approaches (cf. Table 2 in  \cite{kumar2017classification} for a comprehensive review) for this task involve two separate steps of feature extraction and classification. Carneiro et al. \cite{CARNEIRO2017} use histograms and Haralick texture descriptors as an input to a multilayer perceptron. In a similar vein of work, Kumar et al. \cite{kumar2017classification} develop an ensemble of six multilayer perceptron networks trained with GLCM mean features. The approaches by Thomaz et al \cite{thomaz2017feature} and Fonseca et al \cite{fonseca2016breast} apply convolutional neural network  too, however, only to extract features for another classification model (MLP and SVM respectively). Additionally, their results are obtained with a much smaller dataset (less than one thousand exams) of low resolution images. To the best of our knowledge, our work is the first end-to-end four-class breast density classifier using supervised deep convolutional neural network on multi-view mammography images. Importantly, we trained and evaluated our models on a diverse clinically realistic data set of high resolution images, approximately two orders of magnitude larger than any previous work we are aware of.

\section{Conclusions}

In this work, we trained a deep convolutional neural network classifier using a data set of unprecedented size for the task of breast density classification. The level of agreement between the trained classifier and the classes in the data was found to be similar to that between the human experts and the classes in the data, as well as among the human experts themselves. This result strongly suggests that the proposed classifier may have significant clinical relevance, as it provides quantitative, reproducible prediction, while there is often poor intra-reader and inter-reader correlation in the qualitative assessment of breast density tissue, as was observed in our reader study.

\bibliographystyle{IEEEbib}
\bibliography{refs}

\end{document}